\documentclass[10pt,twocolumn,letterpaper]{article}

\usepackage{iccv}
\usepackage{times}
\usepackage{epsfig}
\usepackage{graphicx}
\usepackage{amsmath}
\usepackage{amsthm,amsfonts,amssymb}
\usepackage{pdfsync}
\usepackage{subfig}
\usepackage[font=footnotesize]{caption}
\usepackage{multirow}

\def\mat#1{\mathchoice{\mbox{\boldmath$\displaystyle\tt#1$}}
{\mbox{\boldmath$\textstyle\tt#1$}}
{\mbox{\boldmath$\scriptstyle\tt#1$}}
{\mbox{\boldmath$\scriptscriptstyle\tt#1$}}}

\def\vec#1{\mathchoice{\mbox{\boldmath  $\displaystyle\bf#1$}}
{\mbox{\boldmath  $\textstyle\bf#1$}}
{\mbox{\boldmath  $\scriptstyle\bf#1$}}
{\mbox{\boldmath  $\scriptscriptstyle\bf#1$}}}

\newlength{\colwidth}
\newcommand{\SKIP}[1]{} 

\newcommand{\mbegin} {\left [ \begin{array}}

\newcommand{\mend}   {\end{array} \right ]}

\newcommand{\vbegin} {\left ( \begin{array}{c}}

\newcommand{\vend} {\end{array}\right )}

\def\squareforqed{\hbox{\rlap{$\sqcap$}$\sqcup$}}

\def\qed{\ifmmode\squareforqed\else{\unskip\nobreak\hfil
	\penalty50\hskip1em\null\nobreak\hfil\squareforqed
	\parfillskip=0pt\finalhyphendemerits=0\endgraf}\fi}

\def\vec#1{\mathchoice%
	{\mbox{\boldmath $\displaystyle\bf#1$}}
	{\mbox{\boldmath $\textstyle\bf#1$}}
	{\mbox{\boldmath $\scriptstyle\bf#1$}}
	{\mbox{\boldmath $\scriptscriptstyle\bf#1$}}}

\newcommand{\showeqnlabel}{
	\hbox to 0pt{\quad\quad\relax\fbox{\scriptsize\rm\eqnlblx}%
	\gdef\eqnlblx{xxxx}}} \newcommand{\eqnlblx}{}

\def\@eqnnum{\rm (\theequation)\showeqnlabel}

\newcommand{\nofig}[1]{\centerline{\bf Figure here}}

\def\mat#1{\mathchoice{\mbox{\boldmath$\displaystyle\tt#1$}}
	{\mbox{\boldmath$\textstyle\tt#1$}}
	{\mbox{\boldmath$\scriptstyle\tt#1$}}
	{\mbox{\boldmath$\scriptscriptstyle\tt#1$}}}

\newcommand{\minus}{\scalebox{0.5}[1.0]{$-$}}

\iccvfinalcopy 

\def\iccvPaperID{411} 

\begin{document}

\title{3D Pose from Detections}

\author{Cosimo Rubino
\and
Marco Crocco
\and
Alessandro Perina
\and
Vittorio Murino
\and
Alessio Del Bue \\
\\
Pattern Analysis and Computer Vision Department (PAVIS)\\
Istituto Italiano di Tecnologia\\
Via Morego 30, 16163 Genova, Italy 
}

\maketitle

\begin{abstract}
   We present a novel method to infer, in closed-form, a general 3D spatial occupancy and orientation of a collection of rigid objects given 2D image detections from a sequence of images. In particular, starting from 2D ellipses fitted to bounding boxes, this novel multi-view problem can be reformulated as the estimation of a quadric (ellipsoid) in 3D. We show that an efficient solution exists in the dual-space using a minimum of three views while a solution with two views is possible through the use of regularization. However, this algebraic solution can be negatively affected in the presence of gross inaccuracies in the bounding boxes estimation. To this end, we also propose a robust ellipse fitting algorithm able to improve performance in the presence of errors in the detected objects. Results on synthetic tests and on different real datasets, involving real challenging scenarios, demonstrate the applicability and potential of our method.
\end{abstract}

\section{Introduction} \label{sec:intro}

Localising and finding the pose of objects in generic scenes is a fundamental step for higher level scene understanding. This task is of such importance that major efforts have been put in Computer Vision research in order to obtain efficient detectors from single images. These methods give remarkable results in finding several objects categories in natural scenes and thus putting the basis for the understanding of the visual world. However, while these results are certainly compelling, detection algorithms have been mostly restricted to the estimation of the objects position in 2D. When imaging the same scene from different viewpoints, a crucial question is  how to deal with both 3D geometrical reasoning and higher-level object representation.  

Recent works have clearly pointed out that this lack of 3D reasoning is limiting and that bridging the gap between object detection and multi-view geometry might provide surprising improvements in classical approaches. Starting from the work of Hoeim et al. \cite{Hoiem:etal:2008}, the inclusion of 3D scene reasoning and rules has provided higher detection accuracies. 
Notably, attempts of unifying geometry and object representation have been achieved by defining elaborated Maximum a Posteriori (MAP) inference \cite{bao:savarese:2011} or bundle adjustment with objects \cite{Fioraio:DiStefano:2013}. This way of pursuing high-level object reasoning in multi-view geometry has also inspired novel methods in Simultaneous Localisation and Mapping (SLAM) \cite{salas:etal:2013}. 
Semantic information has been used, on the other hand, to infer the 3D shape of objects and the cameras viewpoints  \cite{vicente2014reconstructing}.
%
Differently, several works attempted to tackle 3D pose estimation from single images only. This severely under-constrained problem has to be solved with the use of strong semantic information (geometrical and physical constraints, i.e. object lying on the ground plane), 2D appearance and shape in the form of 3D wireframe or CAD models \cite{silberman:etal:2012,zia:etal:2013,Zia2014,xiang2012estimating}. Hejrati and Ramanan \cite{hejrati2014analysis} instead use an analysis by synthesis approach that, guided by visual evidence, selects matching HOG patches that best represent a 3D object.

\begin{figure}[t]
\centering
\includegraphics[width=0.90\linewidth]{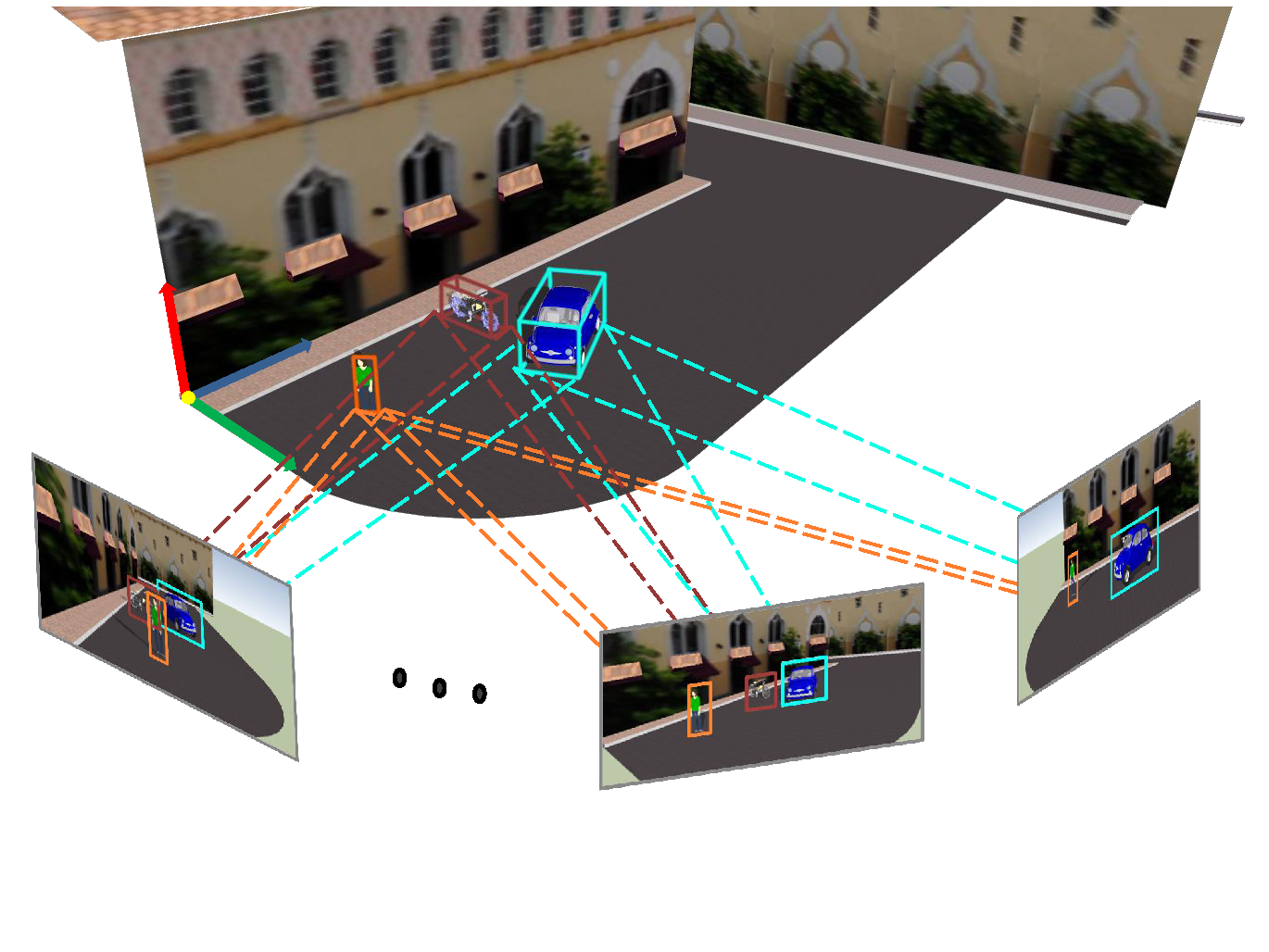}
\vspace{-0.8cm}
  \caption{Example of a set of images of a given 3D rigid scene taken from a camera at different viewpoints. The problem is to recover the 3D occupancy of each object given the 2D BBs detected at each image frame.}
\label{fig:fig1}
\end{figure}

This work takes a different path from previous 3D pose estimation methods by showing that it is possible to obtain accurate 3D object localisations and orientations in closed-form using multi-view relations given only a set of image detections, i.e. 2D bounding boxes (BBs). To the best of the authors knowledge, this work is the first to provide such solution in 3D scene understanding. The solution proposed here recalls standard Structure from Motion (SfM) approaches \cite{Tomasi:Kanade:1992,sturm:triggs:1996} where multiple views of the same rigid scene are used to obtain a 3D reconstruction given a set of 2D feature points measurements. Similarly, we define a novel problem, namely \textit{Pose from Detections (PfD)}, that attempts to estimate the 3D pose (position and orientation) of an object from a set of BBs obtained by general purpose detectors in multiple views (see Fig. \ref{fig:fig1} for a graphical representation). It is worth noting that the occupancy boundary encapsulates the estimation of object center, pose, size and aspect ratio in 3D. Moreover, differently from previous works, we do not need geometric or semantic priors out of 2D bounding boxes from detections, nor advanced detectors yielding pose categorization. 
In particular, we show that there exists an efficient closed-form solution to 3D object localisation for a minimum number of three views if the problem is reformulated as the estimation of a quadric in 3D. This solution is not viable to derive using  BBs (which is a piecewise-defined curve), but is mathematically feasible given a set of ellipses fitted to the original 2D BBs as extracted by the detectors. We also show that for the ill-posed case of two views, it is possible to apply regularisation obtaining performance close to the well-posed solution. 
Moreover, regularisation boosts results whenever the BBs are inaccurate as it might happen in realistic scenarios. Finally, we further introduce a robust segmentation algorithm that can increase the reliability of the 2D ellipse fitting stage by finding the object profile inside the BBs. In such a way, improved results are achieved in challenging real scenarios on different freely available datasets.

The rest of the paper is structured as follows. Section \ref{sec:problem_statement} defines the problem and the related mathematical formalisation. Section \ref{sec:dualsf} presents the PfD solution while Section \ref{sec:fittreg} describes the regularization method and robust ellipse fitting approach. Experiments on real and synthetic data are discussed in Section \ref{sec:exp} and then followed by concluding remarks in Section \ref{sec:concl}.

\section{PfD problem statement} \label{sec:problem_statement}
Let us consider a set of image frames $f=1 \dots F$  representing a 3D scene under different viewpoints. A set of $i=1 \dots N$ rigid objects is placed in arbitrary positions and each object can be detected in each of the $F$ images.  Each object $i$ in each image frame $f$ is identified by a 2D BB $B_{if}$ given by a generic object detector. The BB is defined by a triplet of parameters: $ B_{if} = \left\lbrace  w_{if}, h_{if},\vec c_{if}\right\rbrace$, where $w_{if}$ and $h_{if}$ are two scalars defining the BB height and width respectively and $\vec c_{if}$ is a $2$-vector defining the BB center.

\begin{figure}
	\centering 
	\includegraphics[width=8cm]{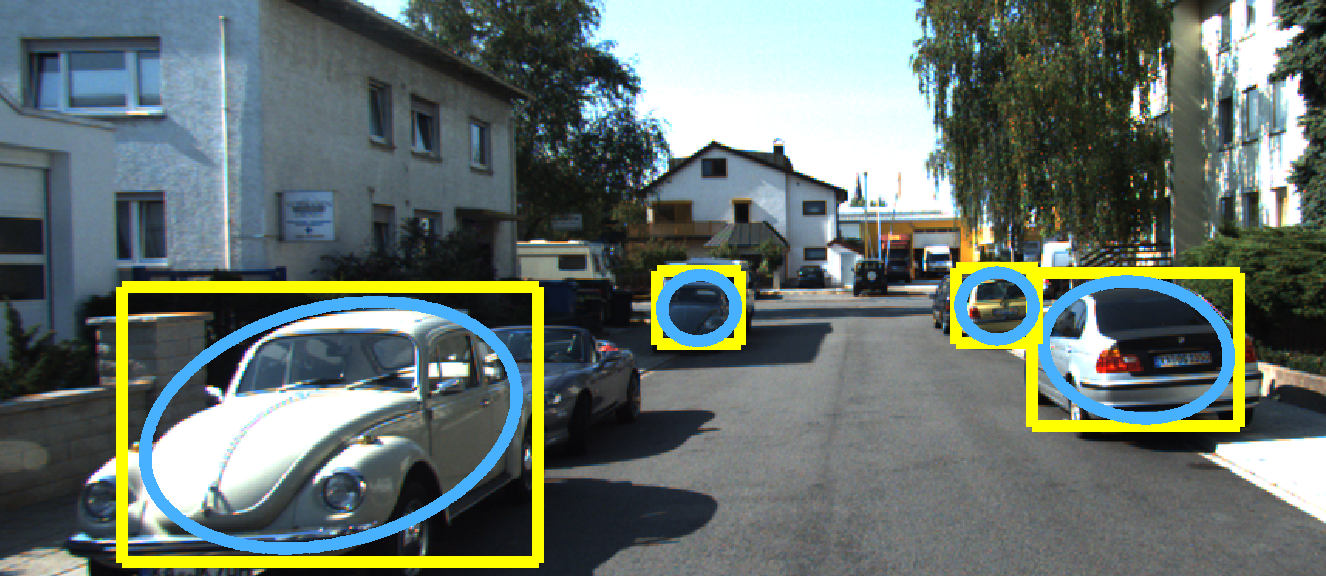}\hspace{.1cm} 
	\caption {Example of BBs (yellow) and corresponding fitted ellipses (blue) for a set of objects.}
	\label{fig:example}
\end{figure}

Our goal is to estimate the pose (position and orientation) of each object in the 3D scene given the 2D BBs using multi-view constraints. 
In order to ease the mathematical formalization of the problem, we move from a BB representation of an object to an ellipsoid one. This is done by associating at each $B_{if}$ an ellipse fitting $\hat{\mat C}_{if}$ that inscribes the BB, as shown in Fig. \ref{fig:example}. 

The aim of our problem is to find the 3D ellipsoids $\mat Q_i$ 
whose projections on the image planes, associated to each frame $f= 1 \dots F$, best fit the 2D ellipses $\hat{\mat C}_{if}$ in the image plane. 
This will solve for both the 3D localisation and occupancy of each object starting from the image detections in the different views.

In the following, we represent each ellipse using the homogeneous quadratic form of a conic equation:
\begin{equation} \label{eq:con}
\vec u^{\top} \hat{\mat{C}}_{if} \,\, \vec u =0,
\end{equation}
where $\vec u \in \mathbb{R}^{3}$ is the homogeneous vector of a generic 2D point belonging, to the conic defined by the symmetric matrix $ \hat{\mat{C}}_{if} \in \mathbb{R}^{3 \times 3}$. 
The conic has five degrees of freedom, given by the six elements of the lower triangular part of the symmetric matrix $\hat{\mat{C}}_{if}$ except one for the scale, since Eq. (\ref{eq:con}) is homogeneous in $\vec{u}$ \cite{hartley2003multiple}. 
Similarly to the ellipses, we represent the ellipsoids in the 3D space with the homogeneous quadratic form of a quadric equation:
\begin{equation} \label{eq:quad}
\vec x^\top \mat Q_i \, \vec x = 0, 
\end{equation}
where $\vec x \in \mathbb{R}^{4}$ represents an homogeneous 3D point belonging to the quadric defined by the symmetric matrix  $\mat Q_i \in \mathbb{R}^{4 \times 4}$. The quadric has nine degrees of freedom, given by the ten elements of the symmetric matrix $\mat Q_i$ up to one for the overall scale.
 
Each quadric $\mat{Q}_i$, when projected onto the image plane, gives a conic denoted with $\mat{C}_{if} \in \mathbb{R}^{3 \times 3}$.
The relationship between $\mat{Q}_i$ and $\mat{C}_{if}$ is defined by the projection matrices $\mat P_{f} \in \mathbb{R}^{3 \times 4}$ associated to each frame. Such matrices can be estimated from the image sequence using standard self-calibration methods \cite{hartley2003multiple,pollefeys1999self}.

\section{Dual space fitting} \label{sec:dualsf}
%
Since the relationship between $\mat{Q}_i$ and $\mat{C}_{if}$ is not straightforward in the primal space, i.e. the Euclidean space of 3D points (2D points in the images) , it is convenient to reformulate it in dual space, i.e. the space of the planes (lines in the images)  \cite{Cross}. 
In particular, the conics in 2D can be represented by the envelope of all the lines tangent to the conic curve, while the quadrics in 3D can be represented by the envelope of all the planes tangent to the quadric surface. 
Hence, the dual quadric is defined by the matrix  $\mat Q^*_i = adj(\mat Q_i)$, where $adj$ is the adjoint operator, and the dual conic is defined by ${\mat{C}}^*_{if} = adj({\mat C}_{if})$ \cite{hartley2003multiple}. 
Considering that the dual conic ${\mat C}^*_{if}$, like the primal one, is defined up to an overall scale factor $\beta_{if}$, the relation between a dual quadric and its dual conic projections ${\mat C}^*_{if}$ can be written as: 
%
\begin{equation} \label{eq:dualq2c}
\beta_{if} {\mat{C}}^*_{if} =  \mat P_{f}  \mat Q^*_i  \mat P^{\top}_{f}.
\end{equation}
%
In order to recover $\mat Q^{*}_{i}$ in closed form from the set of dual conics $ \lbrace {\mat C}^*_{if} \rbrace_{f=1 \dots F} $, we have to re-arrange Eq. (\ref{eq:dualq2c}) into a linear system. Let us define  $\vec v^{*}_{i} = vech (\mat Q_i^*)$ and ${\vec c}^{*}_{if} =vech ({\mat C}^*_{if})$ as the vectorization of symmetric matrices $\mat Q^{*}_{i}$ and ${\mat C}^*_{if}$ respectively\footnote{The operator $vech$ serializes the elements of the lower triangular part of a symmetric matrix, such that, given a symmetric matrix $\mat X \in \mathbb{R}^{n \times n}$, the vector $\vec x$, defined as $\vec x = vech(\mat X)$, is $\vec x \in \mathbb{R}^{g} $ with $ g=\frac{n(n+1)}{2}$.}. 
Then, let us arrange the products of the elements of $\mat P_{f}$ and $\mat P^{\top}_{f}$ in a unique matrix $\mat G_{f} \in \mathbb R^{6 \times 10}$ as follows \cite{henderson1979vec}:  
\begin{equation}
\mat G_f  = \mat D (\mat P \otimes \mat P) \mat E
\end{equation}
where $\otimes$ is the Kronecker product and matrices $\mat D \in \mathbb{R}^{6 \times 9}$ and $\mat E \in \mathbb{R}^{16 \times 10}$ are two matrices such that $vech (\mat X) =\mat D \hspace{0.1 cm} vec(\mat X)$ and $vec (\mat X) =\mat E \hspace{0.1 cm} vech(\mat X)$ respectively,  where $\mat X$ is a symmetric matrix\footnote{The operator $vec$ serializes all the elements of a generic matrix.}.
Given $\mat G_{f}$, we can rewrite Eq. (\ref{eq:dualq2c}) as:
\begin{equation} \label{eq:eqOut}
\beta_{if} {\vec c}^*_{if} = \mat G_{f} \vec v^{*}_{i}.
\end{equation}
In order to get a unique solution for $\vec v^{*}_i$ at least three image frames are needed. Therefore, stacking column-wise 
Eqs. (\ref{eq:eqOut}) for $f = 1 \dots F$, with $F \geq 3$, we have that:
\begin{equation}\label{eq:eqsys}
\mat M_{i} \vec w_{i} = \vec 0_{6F},
\end{equation}
where the matrix $\mat M_i \in \mathbb{R}^{6F \times (10+F)}$ and the vector $\vec w \in \mathbb{R}^{10+F}$ are defined as follows:
\begin{equation}\label{eq:eqsysmx}
\mat M_{i} = 
\arraycolsep=1.4pt
\begin{bmatrix}
\mat G_1 		& \minus {\vec c}_{i1}^* 	& \vec 0_6 		& \vec 0_6 		& \ldots & \vec 0_6 \\
\mat G_2 	& \vec 0_6   	& \minus {\vec c}_{i2}^{*}	& \vec 0_6  		& \ldots & \vec 0_6 \\
\mat G_3 	& \vec 0_6  	& \vec 0_6 		& \minus {\vec c}_{i3}^{*}  & \ldots & \vec 0_6 \\
\vdots 		& \vec 0_6  	& \vec 0_6 		& \vec 0_6   		& \ddots & \vec 0_6 \\
\mat G_F	& \vec 0_6  	& \vec 0_6 		& \vec 0_6   		& \ldots & \minus {\vec c}_{iF}^{*} \\
\end{bmatrix} \;\;\;
\vec w_{i} = 
\begin{bmatrix}
\vec v_i^*\\
\vec{\beta}_i
\end{bmatrix}
\end{equation}
with 
\begin{equation}
\vec{\beta}_i = \left[ \beta_{i1},\beta_{i2},\cdots,\beta_{iF} \right]^{\top}
\end{equation}
and $\vec 0_{d} $ being a zero column vector of length $d$.

Note that in real cases the ellipses $\hat{\mat C}_{if}$ computed by a general purpose object detector might be inaccurate regarding the location of the BB and the window size. Likewise, this will have an effect on the ellipse fitting, inducing an error on the $\mat C_{if}$.
%
%
For this reason, if $\tilde{\mat M}_i$ is the matrix given by object detections 
we can find the solution by minimizing: 
\begin{equation} \label{eq:argmin}
 \tilde{\vec{w}}_{i}= \arg \min_{\vec w_i} \| \tilde{\mat M}_i \vec w_i \|_2^2 \hspace{0.3cm} s.t. \hspace{0.2cm} \|\vec w_i\|^2_2=1,
\end{equation}
where the equality constraint $\|\vec w_i\|^2_2=1$ avoids the trivial zero solution. 
The solution to the minimization problem in Eq. (\ref{eq:argmin}) can be obtained via $\mat{SVD}$ on the $\tilde{\mat M}_i$ matrix, taking the right singular vector associated to the minimum singular value.
The first $10$ rows of $\tilde{\vec{w}}_i$  are the vectorized elements of the estimated dual quadric denoted by $\tilde{\vec v}^*_i$. To get back the estimated matrix of the quadric in the primal space, we obtain first the dual estimated quadric by $\tilde{\mat{Q}}^*_i = vech^{-1}( {\tilde{\vec v}_i^* })$, and subsequently apply the following relation:
\begin{equation} \label{eq:invAdj}
 \tilde{\mat{Q}}_{i} = |\det{\tilde{\mat{Q}}^*_i}|^{\frac{1}{3}} (\tilde{\mat{Q}}^*_i)^{\minus 1}.
\end{equation}
%
%
\section{Regularisation and robust ellipse fitting} \label{sec:fittreg}
%
%
Image detectors can provide inaccurate results given occlusions, illumination variations and complex object poses. %
To this end, we propose a regularisation approach for Eq. (\ref{eq:argmin}) and a method to re-adjust the object  orientation and size from the BBs extracted in the image sequence.

\subsection{Regularized cost function} \label{subsec:reg}
As seen in the previous section, moving to the dual space allows an efficient linearisation of the problem. However this implies a drawback: the algebraic minimization is carried on the dual quadrics in Eq. (\ref{eq:argmin}) and the primal one is obtained by a matrix inversion as in Eq. (\ref{eq:invAdj}). In the presence of noise the matrix $\tilde{\mat{Q}}^*_i$ may become ill-conditioned and therefore small errors may cause relatively large errors in the estimated primal quadric. This problem is particularly evident with the ill-posed case of two views, where the solution to Eq. \ref{eq:eqsys} is a 1-dimensional family of quadrics (i.e. is not unique), or whenever few camera views are spanning a limited range of angles. In such a case, the estimated quadric may result in a nearly degenerate ellipsoid, with dramatically unbalanced axes lengths, or even in wrong quadrics such as an hyperboloid. 

To tackle this problem, we propose to add a regularization term that penalizes the departure of the estimated quadric from a sphere of a given size and center. In other words, we include a regularization on the aspect ratios of the objects modeled by the ellipsoids. 
A sphere in the dual space, centered at the origin, can be represented as a $4 \times 4$ diagonal matrix with the first three diagonal elements positive and equal each other and the fourth diagonal element negative. To account for arbitrary translations, it is sufficient  to pre- and post-multiply the diagonal matrix by a translation matrix. Hence a generic sphere in the dual space can be written as a function of a vector of five parameters  $\vec s=\left[ t_1,t_2,t_3,a,b \right]^{\top}$ as: 
\begin{equation}
\mat S^*(\vec s) = \mat T \mat D \mat T^{\top}
\end{equation}
with 
\begin{equation}
\mat T = 
\begin{bmatrix}
1 & 0 & 0 & t_1\\
0 & 1 & 0 & t_2\\
0 & 0 & 1 & t_3\\
0 & 0 & 0 & 1\\
\end{bmatrix},
\hspace{.6cm}
\mat D=
\begin{bmatrix}
a & 0 & 0 & 0\\
0 & a & 0 & 0\\
0 & 0 & a & 0\\
0 & 0 & 0 & \minus b\\
\end{bmatrix},
\label{D}
\end{equation}
with $a>0$ and  $b>0$.
Thus, the regularization term can be defined as: 
\begin{equation}
R\left(\hat{\vec{v}}^*_i,\vec s \right) = \|  \hat{\vec{v}}^*_i -vech \left(\mat S^*\left(\vec s \right)\right)\|^2_2
\label{reg}
\end{equation}
where $\hat{\vec{v}}^*_i$ is the vectorized $i$-th dual quadric normalized by minus its last (tenth) element giving: 
\begin{equation}
\hat{\vec{v}}^*_i= \vec{v}^*_i/\left(-\vec v_i^*(10)\right).
\label{normal}
\end{equation}
Such normalization avoids the trivial solution when minimizing for Eq. (\ref{reg}). The minus sign has been set in order to preserve the coherence with the constraint $b>0$.\\
Finally, the solution to the regularized problem can be found solving for:  
\begin{equation}
\tilde{\vec f}= \arg \min_{\vec f}\|\tilde{\mat M}_i \hat{\vec w}_i \|^2_2+\lambda R\left(\hat{\vec{v}}^*_i,\vec s \right) \hspace{0.3 cm} s.t. \hspace{0.2 cm} a >0, \hspace{0.1 cm} b > 0
\label{reg_cost}
\end{equation}
where 
\begin{equation}
\vec f = \left[ \hat{\vec w}_i^{\top} \hspace{0.2 cm}  \vec s^{\top} \right]^{\top},  \hspace{1cm} \hat{\vec w}_i = \left[\hat{\vec v}_i^{*\top} \hspace{0.2 cm} \vec{\beta}_i^{\top} \right]^{\top}.
\end{equation}
%
%
%
Note that the normalization in Eq. (\ref{normal}) avoids the need for the quadratic equality constraint $\|\vec w_i\|^2_2=1$ in Eq. (\ref{reg_cost}). 
Finally, the cost function in Eq. (\ref{reg_cost}) is minimized with a nonlinear least squares procedure. 
\paragraph{Initialisation.} Since the convexity of the cost function cannot be guaranteed, a good initialization is mandatory, in particular for the elements of the dual quadric $\hat{\vec{v}}^*_i$. 


First of all we evaluated the sphere with the same center and the same volume of the quadric represented by $\tilde{\vec{v}}^*_i$. Next, we initialized both $\hat{\vec{v}}^*_i$ and the parameters $\vec s$ according to such a sphere.
In detail, let us denote the starting guess values as $a^{(0)}$ and $b^{(0)}$ given by: 
\begin{equation}
a^{(0)}= {|e_{i1}e_{i2}e_{i3}|}^{-1/3}, \hspace{1cm}  b^{(0)}=1,
\label{alpha0}
\end{equation}
where $e_{i1}$, $e_{i2}$ and $e_{i3}$ are the three elements of a vector $\vec e_i$ defined as:
\begin{equation}
\vec e_i=\frac{-det(\tilde{\mat Q}_i)}{det(\tilde{\mat Q}_{i,3\times 3})}eig(\tilde{\mat Q}_{i,3\times 3}) 
\end{equation}
where $eig()$ is the operator that computes the eigenvalues of a square matrix and $\tilde{\mat Q}_{i,3\times 3}$ is the $3 \times 3$ principal submatrix of the primal quadric $\tilde{\mat Q}_i$ found by Eq. (\ref{eq:invAdj})\footnote{
We recall that the volume of an ellipsoid is proportional to $\sqrt{\frac{1}{|e_{i1}e_{i2}e_{i3}|}}$ and the volume of a sphere is proportional to $\sqrt{a^3}$.}. If the initial $\tilde{\mat Q}_i$ is not an ellipsoid, obviously the concept of volume preservation does not hold any more. However the initialization strategy, thanks to the modulus in Eq. (\ref{alpha0}) guarantees to start from a feasible solution corresponding to a sphere.
The initialized translation terms  $t_{1}^{(0)}$ $t_{2}^{(0)}$ and $t_{3}^{(0)}$ are set equal to the three translation parameters extracted by $\tilde{\mat Q}_i$. 

The initial vector $\vec{v}_{i}^{*^{(0)}}$ 
is set equal to $vech(\mat T^{(0)} \mat D^{(0)} \mat T^{(0)^{\top}})$, where $\mat T^{(0)}$ and $\mat D^{(0)}$ are defined by substituting $a^{(0)}$, $b^{(0)}$, $t_{1}^{(0)}$, $t_{2}^{(0)}$ and $t_{3}^{(0)}$ to the corresponding variables in Eq. (\ref{D}). Finally  $\vec{\beta}_{i}^{(0)}=\tilde{\vec{\beta}}_i/\left(-\tilde{\vec v}_i^*(10)\right)$, where $\tilde{\vec{\beta}}_i$ is the vector of scale factors corresponding to the elements of $\tilde{\vec w}_i$ from the $11$-th to the last. 

\subsection{Ellipse fitting from image segmentation}
\label{subsec:fitt}
\begin{figure}[h]
	\centering 
	\includegraphics[width=3.9cm]{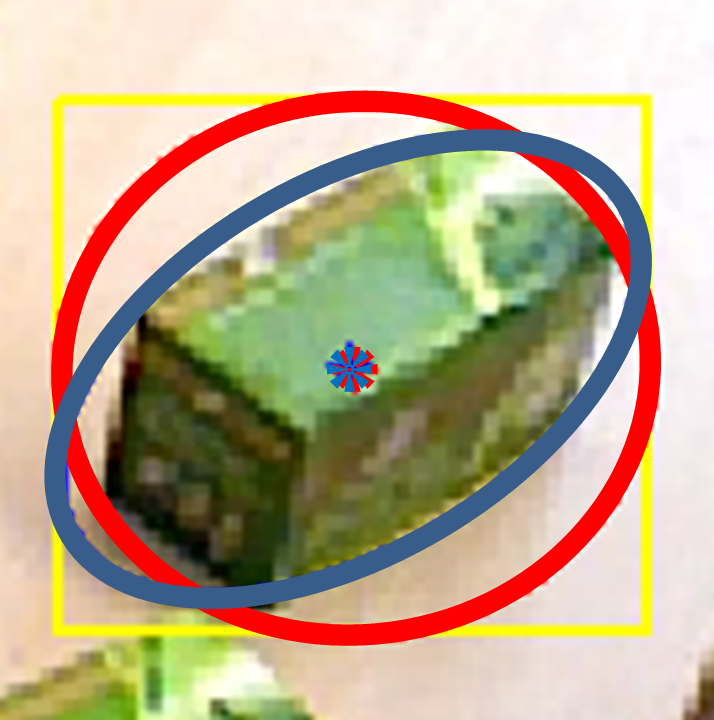}\hspace{.2cm}
    \includegraphics[width=3.9cm]{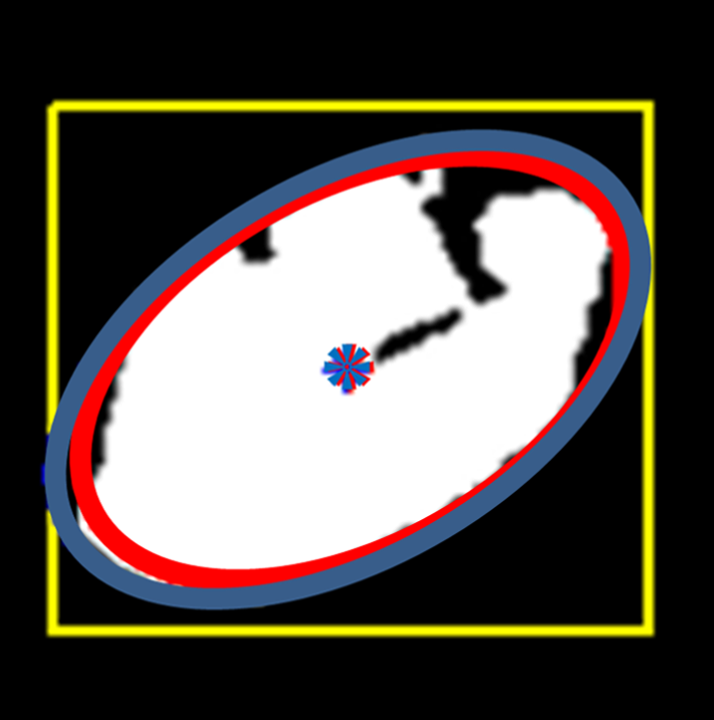}\\
  	 (a)  \hspace{3.6cm} (b) 
	\caption{(a) Example of mismatch between GT ellipse (blue) and ellipse from BB (red); (b) segmented object using SCA algorithm with ellipse fit to the segmentation (red) and GT ellipse (blue). }
	\label{fig:seg}
\end{figure}

%
In general, the BBs from generic object detectors are not precisely aligned with the true object center and often they include a relevant portion of background.
More importantly, BB axes are aligned by construction to the image axes. Thus, the related ellipses are aligned to these axes as well: this results in a relevant rotation mismatch between the real ellipse enclosing the object in the image and the one given by the detector. 
This situation is even more complex when the Ground Truth (GT) ellipses have both strong axes rotation and eccentricity: the GT ellipses are badly approximated by the ellipse obtained from the BB, both in terms of rotation, shape and area (as in Fig. \ref{fig:seg}(a)).

 To cope with these limitations, we design an approach to obtain a tighter and more reliable ellipse fitting on the object inside the BB. 
%
In particular, we adopt a variation of the method proposed in \cite{jojic2009stel}, based on spatial correlations among different image patches, to produce a binary segmentation mask. Here we exploit the fact that the pixel color and intensity distribution of the object in a set of frames are often sufficiently different from the background to allow an accurate object segmentation. Moreover as images come from a video sequence, the pose of the objects smoothly changes from one frame to another and these are the optimal conditions for stel component analysis - SCA algorithm \cite{jojic2009stel}.
SCA is a co-segmentation algorithm that segments images in $S$ segments by learning $K$  components of the model. Such components represent peculiar poses of an object and they are blended together to create a segmentation prior that well adapts to the different poses of the object in the image. The final segmentation is obtained by applying this flexible prior to a GMM-based segmentation.

Given the binary mask localising the object in each BB, we extract the object orientation and size from the covariance matrix $\mat{Cov}_{if}$ of the 2D spatial distribution of the pixels belonging to the object $i$ in frame $f$.
%
Now we can fit a 2D ellipse with axes aligned along the two eigenvectors of $\mat{Cov}_{if}$ and with their magnitude proportional to the square root of its eigenvalues (i.e. $\lambda^j_{if}$ with $j = 1,2$). The magnitude of the ellipse's axes is set making the second central moments of the region enclosed by the ellipse equal to the second 
central moments of the segmented region. An example of this segmentation process is displayed in Fig. \ref{fig:seg}(b) where the matching between GT and fitted ellipses is by far improved with respect to the ellipse from BB in Fig. \ref{fig:seg}(a).

\section{Experiments} \label{sec:exp}
\vspace{-1ex}
The proposed method has been tested on a synthetic scenario and on three real datasets. 
In all the tests, the accuracy of the estimated pose was measured by the volume overlap between GT and estimated ellipsoids ($O_{3D}$) defined as:
\begin{equation}
\label{metric3D}
O_{3D} = \frac{1}{N}\sum_{i=1}^N \frac{\mathcal{Q}_i \cap \tilde{\mathcal{Q}}_i}{\mathcal{Q}_i \cup \tilde{\mathcal{Q}}_i},  
\end{equation}
where $\mathcal{Q}_i$ and $\tilde{\mathcal{Q}}_i$ denote the volume 
of the ellipsoids given by the matrices $\mat Q_i$ and $\tilde{\mat Q}_i$ respectively. 
When the $i$-th estimated quadric is not an ellipsoid $\tilde{\mathcal{Q}}_i$ is set to zero. 
Notice that, when camera views are restricted to a small range of angles or they are related to quasi-planar trajectories, $O_{3D}$ could give poor results even with a small algebraic error in Eq. (\ref{eq:argmin}). However, we chose such metric since it measures in a direct way the success of the algorithm in recovering the 3D position and occupancy of an object.
Moreover, we also evaluated the orientation error by using the measure $ \theta_{err}$, which is the angle in radians between the main axes of estimated and GT ellipsoids. 
In the following we will denote the non regularized solution obtained with SVD (Sec. \ref{sec:dualsf}) as PfD, the regularized solution  (Sec. \ref{subsec:reg}) as PfD+REG, the solution obtained from ellipses fitted to segmented images (Sec. \ref{subsec:fitt}) as PfD+ES and finally the solution implying both regularization and image segmentation as PfD+ES+REG. Regarding orientation estimation, we have reported the SVD solution only since the regularization provides improvement mainly on the 3D localization rather than the object orientation.

\subsection{Synthetic setup}
The synthetic environment contains $50$ ellipsoids randomly placed inside a cube of side $20$ units.
The length of the largest axis $L$ ranges from $3$ to $12$ units, according to a uniform PDF. The lengths of the other two axes are equal to $\gamma L$ with $\gamma \in [0.3,1]$. Finally, axes orientation was randomly generated. 
A set of $20$ views were generated with a camera distance from the cube center of $200$ units and a trajectory was computed so that azimuth and elevation angles span the range $[0^{\circ},60^{\circ}]$ and $[0^{\circ},70^{\circ}]$ respectively\footnote{Notice that the variation of object appearance due to such range of angle views can be handled by state-of-art object detectors \cite{divvala2012important}}. 
Given the projection matrix $\mat P_f$ of each camera frame, GT ellipses given by the exact projections of the ellipsoids were calculated.

Synthetic tests were aimed at validating the robustness of the proposed method against common inaccuracies affecting object detectors, such as coarse estimation of the object center, tightness of the BB with respect to the object size and variations over the object pose.
Thus, each ellipse 
was corrupted by three errors, namely translation error (TE), rotation error (RE) and size error (SE), and fed to the proposed algorithm. 
To impose such errors, the ellipses centers coordinates $c_1$, $c_2$, the axes length $l_1$, $l_2$ and the orientation $\alpha$ of the first axis 
were perturbed 
as follows \footnote{We omit for simplicity the object and frame indexes.}: 
\begin{equation}
\label{trasl}
\hat{c}_j = c_j +\bar{l}{\nu}_j^{\vec c}, \hspace{0.6cm} \hat{\alpha} =  \alpha+\nu^{\alpha}, \hspace{0.6cm} \hat{l}_j = l_j\left( 1+\nu^{l}\right),
\end{equation}
where $\nu_j^{\vec{c}}$, $\nu^{\alpha}$ and $\nu^{l}$ are random variables with uniform PDF and mean value equal to zero, and $\bar{l}=(l_1+l_2)/2$.  
In order to highlight the specific impact of each error,
they were applied separately. Error magnitudes were set tuning the boundary values of the uniform PDFs of $\nu_j^{\vec c}$, $\nu^{\alpha}$ and $\nu^l$.
%
%
\begin{figure*}[thbp!]  
\centering
\includegraphics[width=5cm]{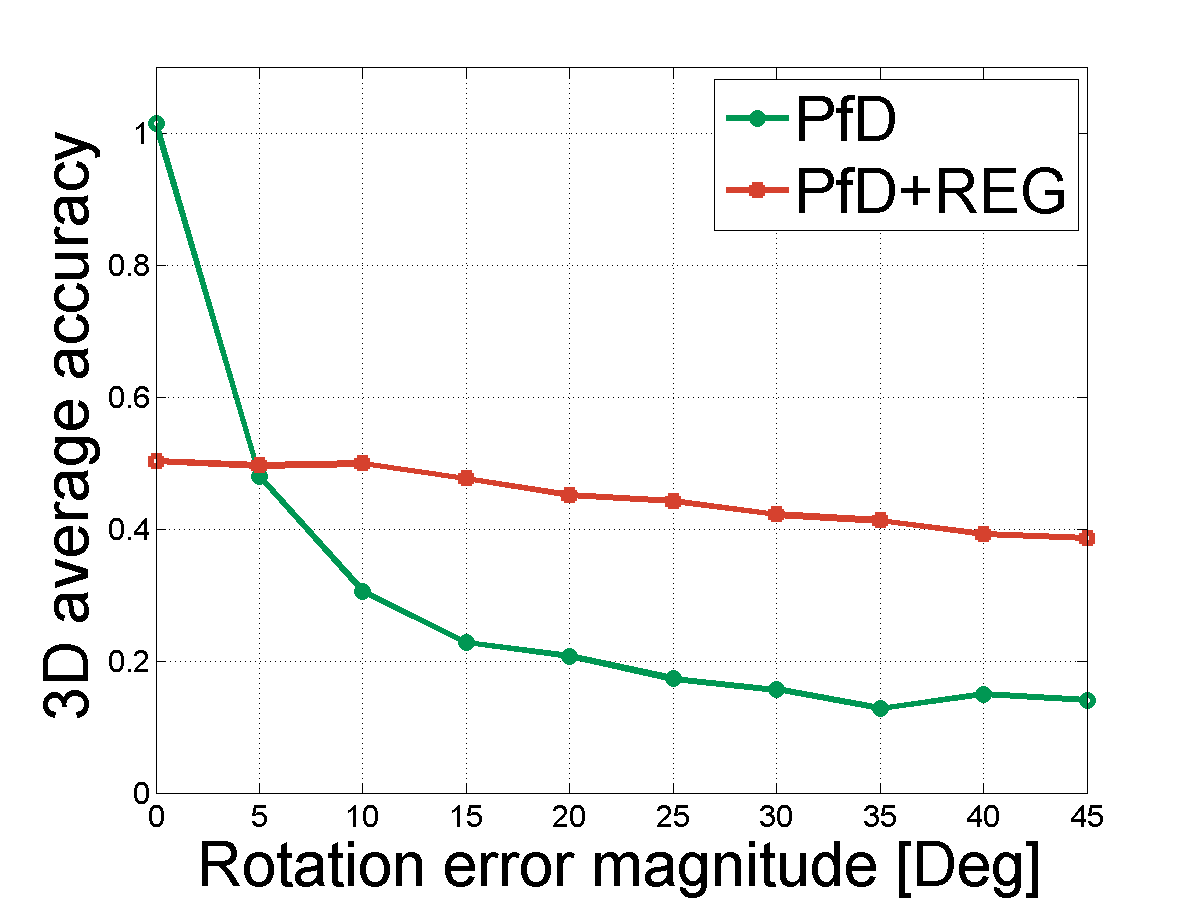}   
\includegraphics[width=5cm]{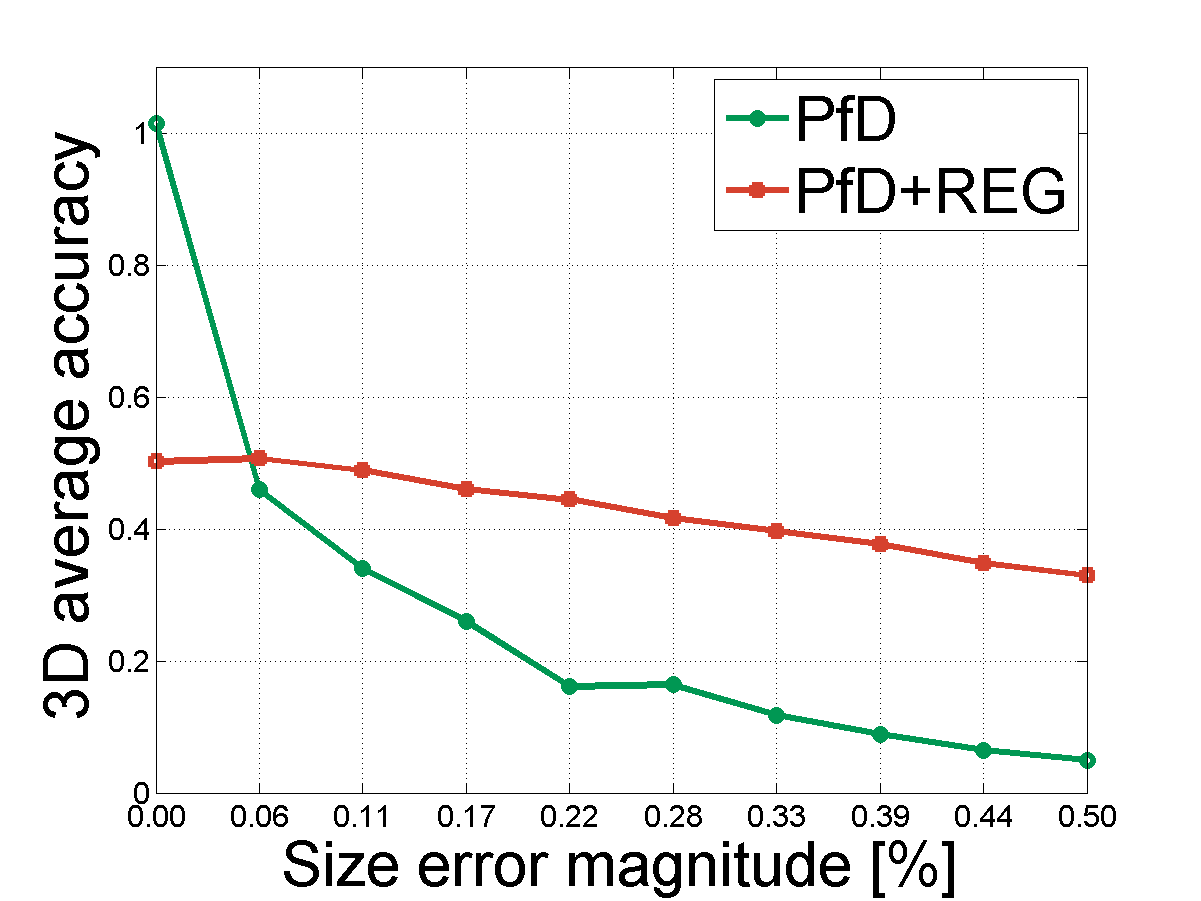} 
\includegraphics[width=5cm]{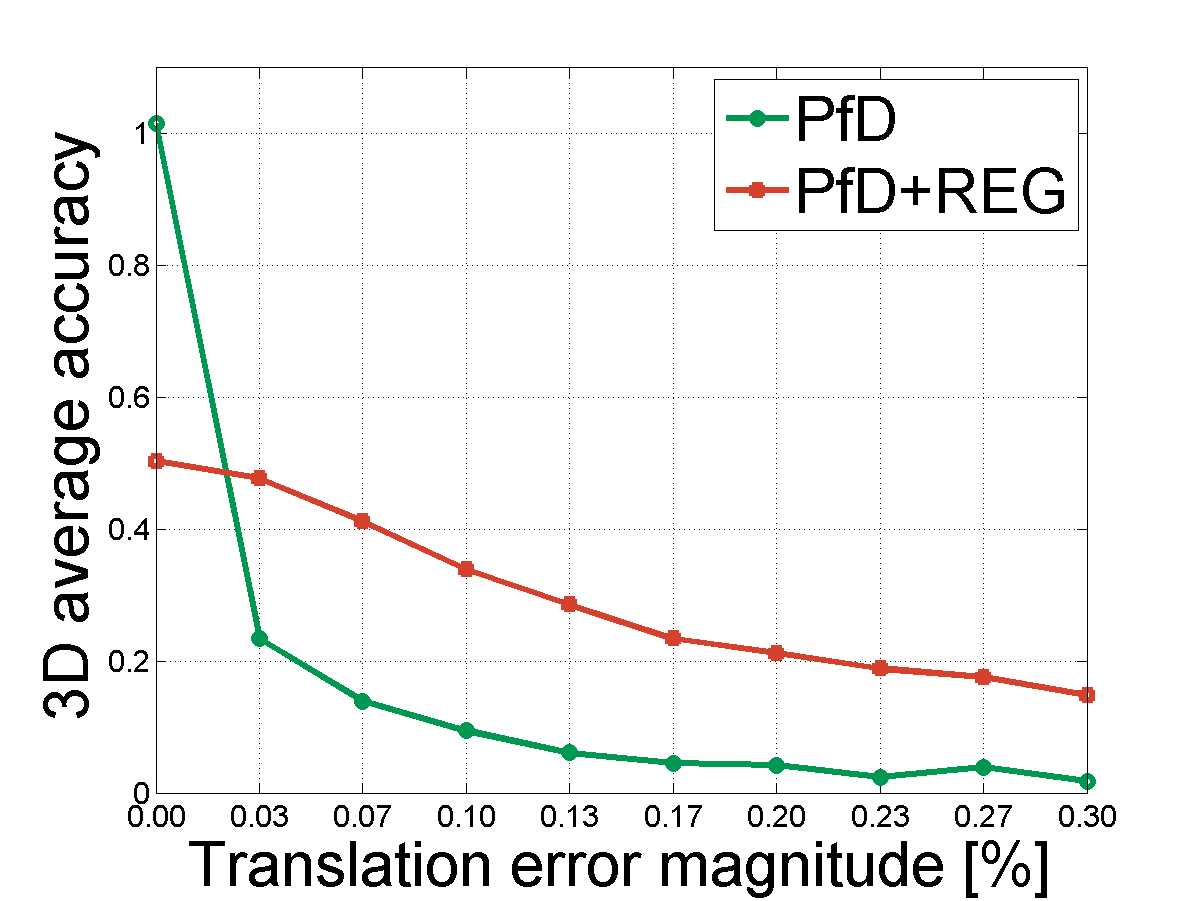} \\
(a) \hspace{4.6cm} (b) \hspace{4.6cm} (c) \\
    \caption{Results for the synthetic tests versus different types of errors. Average accuracy given by PfD or PfD+REG for rotation error (a), size error (b), translation error (c), measured by $O_{3D}$ metric.} 
\label{fig:synth}
\end{figure*}
In Fig. \ref{fig:synth} the average accuracy $O_{3D}$ of the proposed algorithm, for both PfD and PfD+REG, is displayed versus the error magnitude (i.e. the boundary value of the uniform PDF), for RE (Fig. \ref{fig:synth}(a)), SE (Fig. \ref{fig:synth}(b)) and TE (Fig. \ref{fig:synth}(c)). Concerning PfD+REG, the  weighting parameter $\lambda$ has been kept equal for all the tests. 
The results for PfD are perfect in absence of errors but undergo a significant drop as the error increases. This confirms an expected sensitivity of the closed form solution to mismatches between fitted and GT ellipses. Nevertheless, when errors are moderate, the accuracy achieved is remarkable. 

On the contrary, PfD+REG exhibits a very smooth decrease in accuracy versus the errors magnitudes, and a definitely higher performance in comparison to PfD for almost all the error magnitudes, except for the case of zero error (RE, Se and TE) and the $5^{\circ}$ RE. In detail, $O_{3D}$ for PfD+REG, drops from $0.52$ (no errors) to $0.39$, $0.33$ and $0.16$ for maximum RE, SE and TE respectively, while the accuracy of PfD drops from $1$ to $0.15$ (RE), $0.05$ (SE) and $0.02$ (TE) respectively.  


Notice that both the methods are generally more robust toward RE and SE (Figs. \ref{fig:synth}(a), (b) ), than toward TE.
The higher robustness toward RE and SE is quite important since such kind of errors are likely to happen very frequently whenever ellipses are fitted to BBs. Even if the detector is accurate, the BB quantizes the object alignment at steps of $90^{\circ}$, yielding a maximum RE of $45^{\circ}$. This tends to overestimate the object area, thus affecting SE, whenever the object is not aligned to the BB axes (see Fig. \ref{fig:seg}(a)). 
Concerning TE, we noticed that the robustness to such kind of errors decreases when ellipses are far from the image center and small in comparison to the image size. This is mainly due to the structure of the $3 \times 3$ matrix of the conic, whose entries become strongly unbalanced whenever the translation terms in the third row and column increase in respect to the $2 \times 2$ principal sub-matrix, leading likely to numerical problems. However, the maximum degree of failure for the regularized solution is reasonable considering that an object detector placing the BB with a TE of $0.3$ is considered to fail the detection\footnote{The TE is applied independently to both the horizontal and vertical components of the ellipse center, i.e. see Eq. (\ref{trasl}), resulting in a maximum overall translation of $0.3\sqrt{2}$.}. 


%
%
\subsection{ACCV dataset evaluation}
\label{Hint}
%
The ACCV dataset \cite{hinterstoisser2012accv} contains $15$ sequences, each related to a single object laying on a table at different camera viewpoints (from $100$ to $1000$ per sequence). 
We selected the subset of $8$ sequences for which the 3D point cloud of the object is provided, and limit the number of views to $100$ for each sequence.
For each object we evaluated the GT ellipsoid as the envelope of the 3D point cloud. Moreover, for each frame and each object, we generated a 2D BB by simulating the output of a multi-scale object detector providing BBs with variable aspect ratio, like the well known Deformable Part Model (DPM) \cite{Felzenszwalb2010}.

\begin{figure*}[btph]
\centering
\includegraphics[width=4 cm]{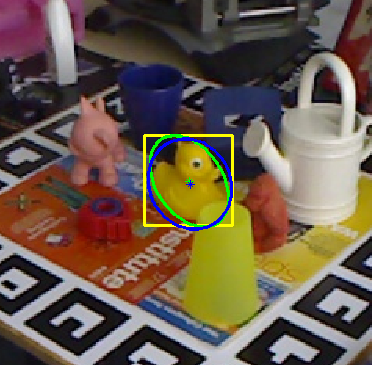} \hspace{.2cm}
\includegraphics[width=4 cm]{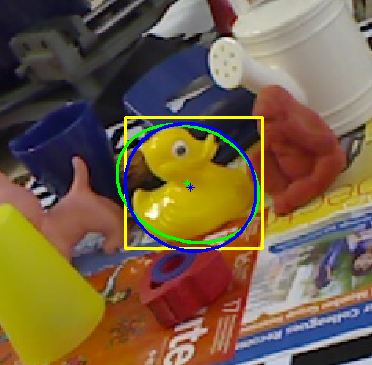} \hspace{.2cm}
\includegraphics[width=4 cm]{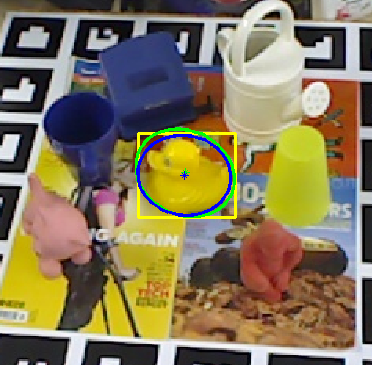} \hspace{.2cm}
\includegraphics[width=4 cm]{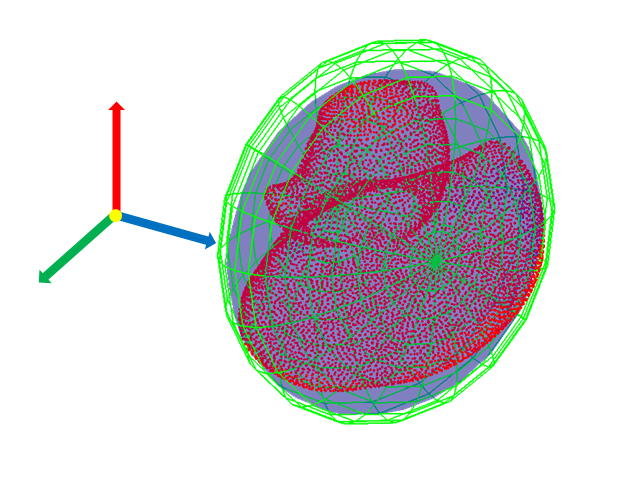} \\ \vspace{.3cm}
\includegraphics[width=4 cm]{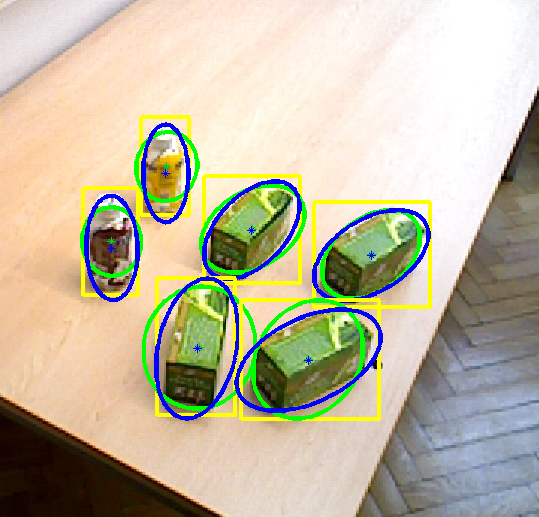} \hspace{.2cm}
\includegraphics[width=4 cm]{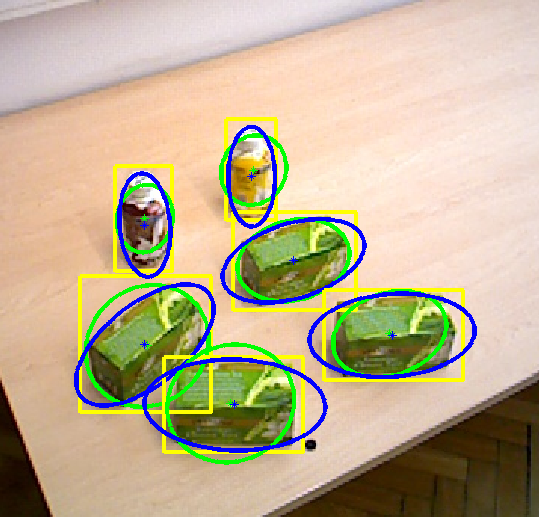} \hspace{.2cm}
\includegraphics[width=4 cm]{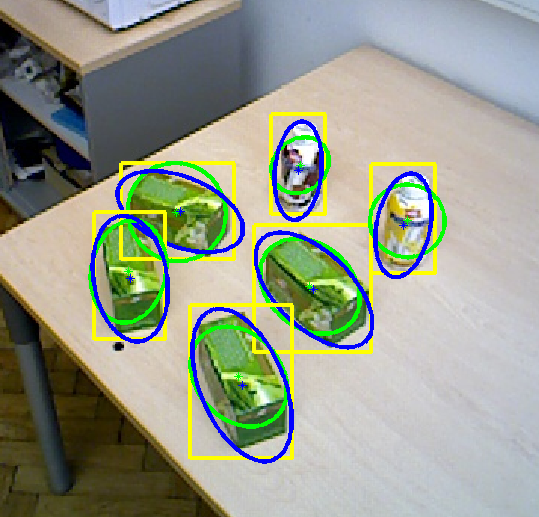} \hspace{.2cm}
\includegraphics[width=4 cm]{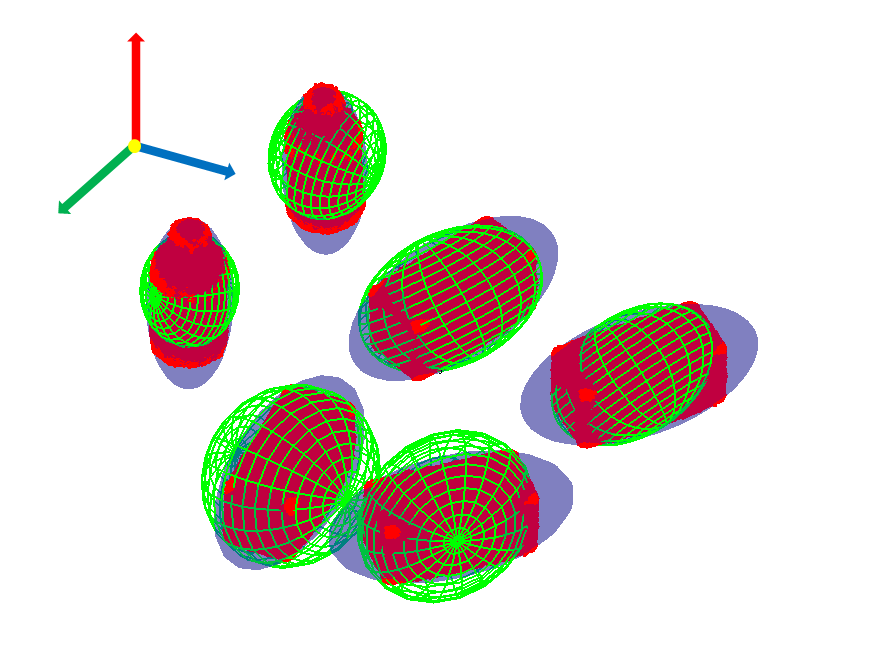} \\
\caption{Localisation results for the ACCV (first row), TUW (second row) datasets. The first three columns show a close up of the views with the output of a generic object detector (yellow BB) and projections of GT and reconstructed ellipsoids (blue and green ellipses respectively). the last column show the localisation of the object (red), GT ellipsoid (blue) and estimated ellipsoid (green).}
\label{fig*:real_exp}
\end{figure*}

In Fig. \ref{fig*:real_exp}, first row, an example of the localisation performance for the PfD method is displayed for the "duck" sequence. The estimated ellipsoid almost perfectly fits the GT one in respect to location, size, eccentricity and alignment, as it can be seen in three frames from the sequence and in the overall 3D localisation.
In Table \ref{tab:accv} the accuracy for each sequence, for PfD, is reported in terms of $O_{3D}$ and $\theta_{err}$.   
In general, the accuracy is lower for objects with a strongly not convex shape, like the driller, and higher for ellipsoid-like objects like  the duck. 

Due to the large number of views and to the centering of each object in each frame (see discussion on synthetic results), the accuracy  is on average quite good even without regularization and ellipses from BBs, with an $O_{3D}$ of $0.60$. 
For this reason, either adding regularization or fitting ellipses to segmented images did not improve significantly the results on this dataset. 

\begin{figure}[btph]
\centering
\includegraphics[width=8cm]{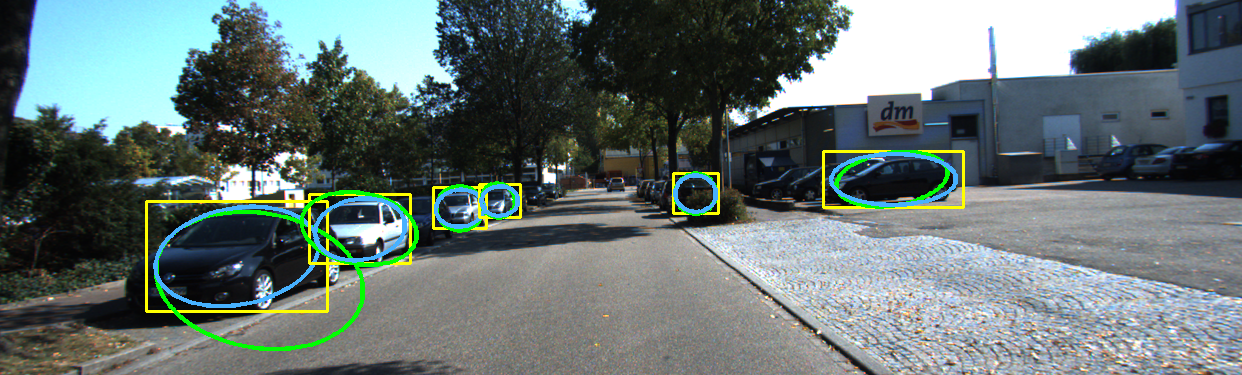} \\ \vspace{.3cm}
\includegraphics[width=8cm]{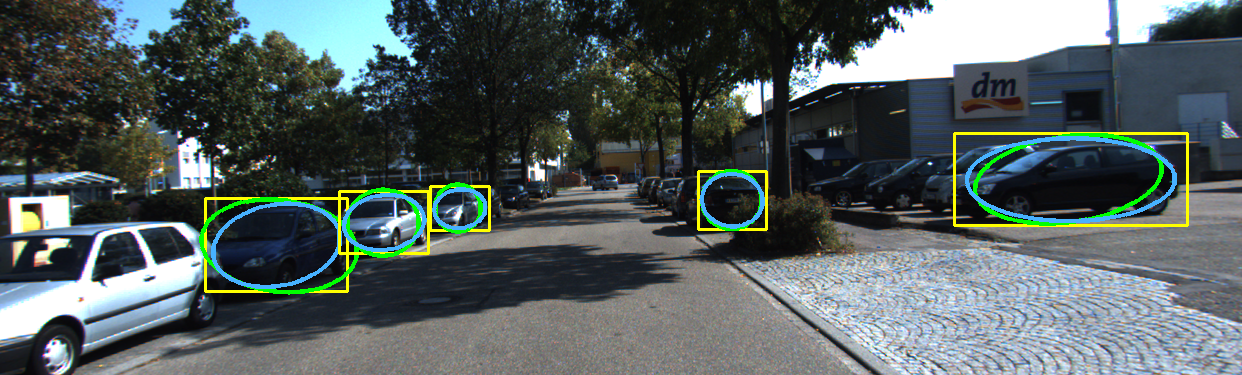} \\ \vspace{.3cm}
\includegraphics[width=8cm]{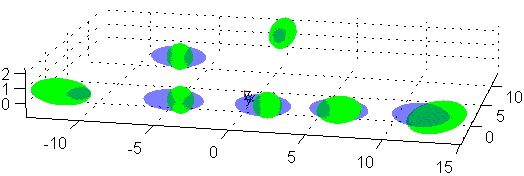}
\caption{Localisation results for KITTI (third row) datasets. The first two images show a close up of the views with the output of a generic object detector (yellow BB) and projections of GT and reconstructed ellipsoids (blue and green ellipses respectively). the last column show the localisation of the object (red), GT ellipsoid (blue) and estimated ellipsoid (green).}
\label{fig*:real_exp2}
\end{figure}

\begin{table}[h]
\caption{$O_{3D}$ and $\theta_{err}$ for sequences from ACCV dataset for PfD.}
\centering
\footnotesize
\resizebox{\columnwidth}{!}{
\begin{tabular}{|c|c|c|c|c|c|c|c|c|c|}
\hline 
& Iron & Duck & Ape & Can & Driller & Vise & Glue & Cat & Avg \\
\hline  
$O_{3D}$ & 0.71 & 0.83 & 0.47 & 0.74 & 0.54 & 0.67 & 0.33 & 0.56 & \bf{0.60}\\ \hline
$\theta_{err}$ & 0.17 &  0.13 &  0.23 & 0.13 & 1.00 & 0.43 & 0.01 &  0.57 & $\bf{ 0.33}$\\
 \hline
\end{tabular}
}
\label{tab:accv}
\end{table}

%

%

\subsection{TUW dataset evaluation}
\label{TUW}
The TUW dataset \cite{aldoma2014automation} has $15$ annotated sequences showing a table with different sets of objects deployed over it. The number of frames per sequence ranges from $6$ to $20$. A 3D point cloud for each object is also provided. As for the ACCV dataset, we obtained the GT ellipsoids for each object and the 2D BBs.
%
We discarded sequences with strong occlusions that cannot be handled by current object detectors, and sequences where objects appear for a number of frames lower than $3$, thus retaining $5$ sequences. 
In Fig. \ref{fig*:real_exp}, second row, an example of the localisation performance with the PfD+ES+REG method for a selected sequence is displayed. The accuracy in the estimation of the objects' pose is remarkable and this trend is confirmed for all the other objects in the ACCV dataset in term of size, eccentricity and alignment of GT ellipsoids. 

All the selected sequences have been tested with the four methods PfD, PfD+REG, PfD+ES and PfD+ES+REG. The accuracy for each sequence is reported in Table \ref{tab:iros}, 
according to $O_{3D}$ and $\theta_{err}$.  
It can be noticed that the regularization yields a sharp improvement in the accuracy, in terms of $O_{3D}$, with respect to non regularized methods on all the tested sequences. In particular, the accuracy is 
even doubled 
moving from PfD to PfD+REG. The ellipse fitting from segmentation yields a further improvement, leading a $5\%$ increment in accuracy moving from PfD+REG to PfD+ES+REG 
and a $3\%$ increment moving from PfD to PfD+ES. 
Remarkably, the improvement from PfD+REG to PfD+ES+REG is achieved in every sequence. 

\begin{table}[h]
\footnotesize
\caption{$O_{3D}$ and $\theta_{err}$ for the sequences from TUW dataset.}
\centering
\tabcolsep=0.08cm
\begin{tabular}{|c|c|c|c|c|c|c|c|}
\hline                                         
              &   & Seq.1 & Seq.7 & Seq.8 & Seq.10 & Seq11 & \textbf{Avg}\\ \hline
\multirow{2}{*}{PfD}& $O_{3D}$  & 0.09 & 0.43  & 0.00 & 0.60 & 0.16 & 0.25 \\ 
              & $\theta_{err}$  & $ 0.69$ & $ 0.73$  & $ NN $ & $ 0.97$ & $ 1.34$  & $ 0.93$ \\ \hline
\multirow{2}{*}{PfD+ES}& $O_{3D}$ & 0.12 & 0.38 & 0.16 & 0.65 & 0.09 & 0.28 \\ 
         & $\theta_{err}$  & $ 0.69$ & $ 0.86$ & $ 0.77$ & $ 0.68$   & $ 1.1$  & $ 0.82$ \\ \hline
PfD+REG   &   $O_{3D}$   & 0.48  & 0.49   & 0.48 & 0.71   & 0.40  & 0.51 \\ \hline
PfD+REG+ES &  $O_{3D}$    & \bf{0.50} & \bf{0.56} & \bf{0.52} & \bf{0.74} & \bf{0.43} & \bf{0.55} \\ \hline

\end{tabular}
\label{tab:iros}
\end{table}

\subsection{KITTI dataset evaluation and comparisons}
\label{KITTI}
The KITTI dataset \cite{Geiger2013} is composed by a set of sequences taken from a camera mounted on a moving car in an urban environment. The dataset provides full annotations for cars appearing in each frame from which GT ellipsoids can be computed. We sampled $6$ sub-sequences displaying parked cars\footnote{The selected sequences (Seq.) and the corresponding frames (Fr.) defining the sub-sequences are the following: Seq 5 (Fr. 142 - 153); Seq. 9 (Fr. 90 - 106); Seq. 22 (Fr. 49 - 86); Seq. 23 (Fr. 1 - 17); Seq. 35 (Fr. 1 - 5); Seq. 36 (Fr. 44 - 57).}.   
We generated 2D BBs simulating a multi-scale object detector for each car and each frame. We pruned out cars with strong occlusions for which a reliable detection is unlikely. The number of remaining cars was of $34$ and the average number of views in which a car is visible is $10$. 
Ellipsoids estimation is particularly challenging on this dataset, since the camera motion is almost planar. 
Moreover, cars are usually placed at the street borders and the camera moves straight in most of the sequences. Hence, the range of angles between car and camera spanned by the sequence of camera views is very narrow and almost limited to the azimuth plane. Finally, each car appears in a limited subset of frames. We did not apply segmentation on this dataset due to the extreme difficulty in segmenting some of the cars that are partially overlapping each other.

In Table \ref{tab:kitti} quantitative results are displayed for the six selected sequences. Despite the difficulty of the dataset, PfD achieves a reasonable result of $O_{3D}=0.17$  and $\theta_{err}= 0.36$ and the use of regularization almost doubles the accuracy, yielding $O_{3D}=0.36$.  
The result for PfD+REG is visually confirmed by looking at Fig. \ref{fig*:real_exp2}: 
all the seven cars are correctly located in the 3D space, with a reasonable precision concerning the object size. The estimated aspect ratio and orientation of the cars is less accurate but acceptable considering the  challenging conditions characterizing this dataset.
Besides the occupancy values $O_{3D}$, our algorithm achieves excellent 3D localization performance. In Table \ref{tab:distances} the percentage of estimated ellipsoid centers within $1$ m and $2$ m from the GT centroids of the objects is reported. With PfD+REG almost all the cars ($93\%$) are within $2$ m and $76\%$ are within $1$ m. Moreover, PfD alone achieves very good results ($83\%< 2$ m and $81\%<1$ m on average). The lower average value of PfD+REG w.r.t. PfD ($<1$ m) can be explained by the fact that regularization enforces the sphere-likeness of the quadric, and due to the non-linearity of the problem, this may sometime decrease the accuracy in the 3D center estimation. Notice that, even for those estimated quadrics not corresponding to feasible ellipsoids, a quadric center can be still computed and also in this case its position is most of the times within $2$ m of the GT centroids.\\
The localization accuracy has been also compared with the one obtained by Choi et al. \cite{choi2010multiple} on the same sequences. Despite the use of priors given by ground plane, results of \cite{choi2010multiple}, displayed in Table \ref{tab:kitti}, are significantly worse than the ones obtained with our method.

\begin{table}[h]
\footnotesize
\caption{$O_{3D}$  and $\theta_{err}$ for the sequences from the KITTI dataset.}
\centering
\begin{tabular}{|p{0.5cm}|p{0.5cm}|c|c|c|c|c|c|c|}
\hline            
         &        	  &  S.5  	& S.9 	& S.22 & S.23	& S.35 & S.36 & \bf{Avg.} \\ \hline
\multirow{2}{*}{PfD}  & $O_{3D}$             &  0.13  	& 0.05 & 0.13  & 0.48 & 0.17  & 0.08  & 0.17 \\  
& $\theta_{err}$ & $0.24$ & $0.22$ & $ 0.16$ & $ 0.21$ & $ 0.55$ & $ 0.07$ & $ 0.24$ \\ \hline
PfD \newline +REG    & $O_{3D}$           &  0.21 	& 0.49  & 0.45 & 0.37   & 0.30 & 0.35 &	0.36 \\ \hline 
\end{tabular}
\label{tab:kitti}
\end{table}


\begin{table}[h]
\caption{Percentages of estimated centroids within 1 m or 2 m w.r.t. GT centroids for the 6 sequences of the KITTI dataset.}
\centering
\footnotesize
\begin{tabular}{|c|c|c|c|c|c|c|c|}
\hline            
                 		& S.5  		& S.9		& S.22	  & S.23	 &  S.35    & S.36   & \bf{Avg}	  \\ \hline
PfD  $<$1m              & $\bf{80}$        & 71        & 86      & $\bf{100}$      & $\bf{50}$       & $\bf{100}$    &  $\bf{81}$    \\ \hline                    	
PfD+REG $<$1m          	& 60        & $\bf{86}$       	& $\bf{100}$	  & 75		 & $\bf{50}$   & 86     &  76     \\ \hline
\cite{choi2010multiple} $<$1m & 20        & 0       	& 10	  & 27		 & 0       & 0     &  10     \\ \hline \hline
PfD $<$2m           	& $\bf{80}$        & 71        & $\bf{100}$      & $\bf{100}$      & 50       & $\bf{100}$   & 83      \\ \hline
PfD+REG $<$2m        	& $\bf{80}$        & $\bf{100}$       & $\bf{100}$     & $\bf{100}$      & $\bf{75}$      & $\bf{100}$   & $\bf{93}$ 	\\ \hline
\cite{choi2010multiple} $<$2m & 40        & 33        & 30      & 45     & 0       & 0   & 25      \\ \hline

\end{tabular}
\label{tab:distances}
\end{table}

\subsection{Minimal configurations test}

In order to stress the method capabilities, we also tested minimal configurations of three and two views, the latter being solvable with regularization only due to its ill-posed nature. We have chosen a set of views with a wide baseline (about $45$ deg in rotation) so making difficult to compute any disparity from the detected objects. Table \ref{tab:Min} shows that in general accuracy is lower (ACCV) or slightly lower (TUW, KITTI) than in case of more views but still remaining reasonable for all the datasets. The results obtained by adding regularisation provides increments and performance in line with the previous experiments for the ACCV and KITTI datasets, while in TUW the third view allows to increase notably the accuracy.



\begin{table}[h]
\footnotesize
\caption{Average $O_{3D}$ for 2 and 3 views for the sequences from ACCV, TUW and KITTI dataset.}
\centering
\tabcolsep=0.08cm
\begin{tabular}{|c|c|c|c|}
\hline                                         
                 & ACCV & TUW & KITTI   \\ \hline
PfD 3views           & 0.46 & 0.20  & 0.12  \\ \hline
PfD+REG 3views       & 0.47     & 0.51  & 0.34  \\ \hline
PfD+REG 2views       & 0.46 & 0.25  & 0.34  \\ \hline
\end{tabular}
\label{tab:Min}
\end{table}

\section{Discussions and Future Work} \label{sec:concl}

This paper presented a closed-form solution to recover the 3D occupancy of objects from 2D detections in multi-view. This algebraic solution is achieved through the estimation of a 3D quadric given 2D ellipsoids fitted at the objects detectors BBs. Moreover a regularization approach was devised to cope with possible ill-conditioning of the problem.  The approach was tested against the common inaccuracies affecting object detectors such as coarse estimation of the object center, tightness of the BB in respect to the object size and variations over the object pose. To strengthen further the approach, a  robust ellipse fitting method was introducing using a segmentation algorithm over the detected BBs in multi-view.   Experiments show that even with relevant errors, the estimated quadrics are able to localise the object in 3D and to define a reasonable occupancy. Moreover, the proposed 
estimation of object orientation, by means of a segmentation algorithm, can be used in order to increase accuracy and the percentage of the reconstructed quadrics.

The solutions of this problem has strong practical breakthroughs given the recent evolution of recognition algorithms. In particular, object detection is certainly going towards increased generality, so providing detectors for several object classes \cite{dean:etal:2013}. Thus, the proposed method can provide a quick and very efficient solution to leverage the 2D information for 3D scene understanding where objects can be inter-related given their position in the metric space. 
This will inject important 3D reasoning in classic frameworks for object detection that are mostly restricted to 2D reasoning.

Regarding future work, mis-detections (i.e. outliers) might affect negatively the estimation of the quadrics. Thus, including further robustness in the optimization, through ad hoc regularization terms in the cost function, might improve the performance of the system. Moreover, the ellipsoid orientations given the BBs can be further improved, especially when objects occlusions or similar appearance textures are present. 

{\small
\bibliographystyle{ieee}
\bibliography{2015_bib_sfom}
}

\end{document}